\documentclass[11pt]{article}
\usepackage{acl2015}
\usepackage{times}
\usepackage{url}
\usepackage{latexsym}
\usepackage{amssymb}
\usepackage{graphicx}
\usepackage{float}
\usepackage{booktabs}
\usepackage{graphics}
\usepackage[fleqn]{amsmath}
\usepackage{multirow}
\title{A Computational Approach to Automatic Prediction of \textit{Drunk-Texting}}
\author{\begin{tabular}{ccc}
Aditya Joshi$^{1,2,3}$ & Abhijit Mishra$^{1}$ & Balamurali AR$^{4}$\\
\end{tabular}\\
\begin{tabular}{ccc}
\multicolumn{3}{c}{\textbf{Pushpak Bhattacharyya$^{1}$ \hspace{1cm} Mark James Carman$^{2}$}}\\
\multicolumn{3}{c}{$^{1}$IIT Bombay, India, $^{2}$Monash University, Australia}\\
\multicolumn{3}{c}{$^{3}$IITB-Monash Research Academy, India $^{4}$Aix-Marseille University, France}\\
\multicolumn{3}{c}{\tt \{adityaj, abhijitmishra, pb\}@cse.iitb.ac.in}\\
\multicolumn{3}{c}{\tt balamurali.ar@lif.univ-mrs.fr,mark.carman@monash.edu}\\
\end{tabular}
}
 \begin{document}
\maketitle
\vspace{-2cm}
\begin{abstract}
Alcohol abuse may lead to unsociable behavior such as crime, drunk driving, or privacy leaks. We introduce automatic drunk-texting prediction as the task of identifying whether a text was written when under the influence of alcohol. We experiment with tweets labeled using hashtags as distant supervision. Our classifiers use a set of N-gram and stylistic features to detect drunk tweets. Our observations present the first quantitative evidence that text contains signals that can be exploited to detect drunk-texting.
\end{abstract}
\section{Introduction}
The ubiquity of communication devices has made social media highly accessible. The content on these media reflects a user's day-to-day activities. This includes content created under the influence of alcohol. In popular culture, this has been referred to as `\textit{drunk-texting}'\footnote{Source: http://www.urbandictionary.com}. In this paper, we introduce automatic `\underline{drunk-texting prediction}' as a computational task. Given a tweet, the goal is to automatically identify if it was written by a drunk user. We refer to tweets written under the influence of alcohol as `\textit{drunk tweets}', and the opposite as `\textit{sober tweets}'. 

A key challenge is to obtain an annotated dataset. We use hashtag-based supervision so that the authors of the tweets mention if they were drunk at the time of posting a tweet. We create three datasets by using different strategies that are related to the use of hashtags. We then present SVM-based classifiers that use N-gram and stylistic features such as capitalisation, spelling errors, etc. Through our experiments, we make subtle points related to: (a) the \textit{performance of our features}, (b) \textit{how our approach compares against human ability} to detect drunk-texting, (c) \textit{most discriminative stylistic features}, and (d) \textit{an error analysis} that points to future work. To the best of our knowledge, this is a first study that shows the feasibility of text-based analysis for drunk-texting prediction.

\section{Motivation}
Past studies show the relation between alcohol abuse and unsociable behaviour such as aggression \cite{aggression}, crime \cite{crime}, suicide attempts \cite{suicide}, drunk driving \cite{driving}, and risky sexual behaviour \cite{condom}. \newcite{suicide} state that ``\textit{those responsible for assessing cases of attempted suicide should be adept at detecting alcohol misuse}''. Thus, a drunk-texting prediction system can be used to identify individuals susceptible to these behaviours, or for investigative purposes after an incident.

Drunk-texting may also cause regret. Mail Goggles\footnote{http://gmailblog.blogspot.in/2008/10/new-in-labs-stop-sending-mail-you-later.html} prompts a user to solve math questions before sending an email on weekend evenings. Some Android applications\footnote{https://play.google.com/store/apps/details?id=com.oopsapp} avoid drunk-texting by blocking outgoing texts at the click of a button. However, to the best of our knowledge, these tools require a user command to begin blocking. An ongoing text-based analysis will be more helpful, especially since it offers a more natural setting by monitoring stream of social media text and not explicitly seeking user input. Thus, automatic drunk-texting prediction will improve systems aimed to avoid regrettable drunk-texting. To the best of our knowledge, ours is the first study that does a quantitative analysis, in terms of prediction of the drunk state by using textual clues. 

Several studies have studied linguistic traits associated with emotion expression and mental health issues, suicidal nature, criminal status, etc. \cite{pennebaker1993putting,pennebaker1997writing}. NLP techniques have been used in the past to address social safety and mental health issues \cite{resnik2013using}.

\section{Definition and Challenges}
Drunk-texting prediction is the task of classifying a text as drunk or sober. For example, a tweet `\textit{Feeling buzzed. Can't remember how the evening went}' must be predicted as `\textit{drunk}', whereas, `\textit{Returned from work late today, the traffic was bad}' must be predicted as `\textit{sober}'. The challenges are:
\begin{enumerate}
\item \textbf{More than topic categorisation}: Drunk-texting prediction is similar to topic categorisation (that is, classification of documents into a set of categories such as `\textit{news}', `\textit{sports}', etc.). However, \newcite{emotionjudgment} show that alcohol abusers have more pronounced emotions, specifically, anger. In this respect, drunk-texting prediction lies at the confluence of topic categorisation and emotion classification.
\item \textbf{Identification of labeled examples}: It is difficult to obtain a set of sober tweets. The ideal label can be possibly given only by the author. For example, whether a tweet such as `\textit{I am feeling lonely tonight}' is a drunk tweet is ambiguous. This is similar to sarcasm expressed as an exaggeration (for example, `\textit{This is the best film ever!}), where the context beyond the text needs to be considered.
\item \textbf{Precision/Recall trade-off}: The goal that a drunk-texting prediction system must chase depends on the application. An application that identifies potential crimes must work with high precision, since the target population to be monitored will be large. On the other hand, when being used to avoid regrettable drunk-texting, a prediction system must produce high recall in order to ensure that a drunk message does not pass through.
\end{enumerate}

\section{Dataset Creation}
\label{datasets}
\begin{table*}[ht!]
\begin{tabular}{|l|p{10cm}|}
\hline
\textbf{Feature} & \textbf{Description} \\ \hline
\multicolumn{2}{|c|}{\textbf{N-gram Features}}                                                   \\ \hline
Unigram \& Bigram (Presence) & Boolean features indicating unigrams and bigrams\\
Unigram \& Bigram (Count)  & Real-valued features indicating unigrams and bigrams\\ \hline
%Unigram (Count)    & Bigram (Count)      & Trigram (Count)         & Unigrams from LDA (Count)    \\ \hline
\multicolumn{2}{|c|}{\textbf{Stylistic Features}}                                                    \\ \hline
LDA unigrams (Presence/Count)    & Boolean \& real-valued features indicating unigrams from LDA\\
POS Ratio        &  Ratios of nouns, adjectives, adverbs in the tweet\\
\#Named Entity Mentions & Number of named entity mentions\\
\#Discourse Connectors      & Number of discourse connectors \\ 
%\multicolumn{4}{|c|}{\textbf{Stylistic Features}}                                                 \\ \hline
Spelling errors   & Boolean feature indicating presence of spelling mistakes\\
Repeated characters     & Boolean feature indicating whether a character is repeated three times consecutively\\
Capitalisation       & Number of capital letters in the tweet\\
%\multicolumn{4}{|c|}{\textbf{Sentiment-based Features}}                                           \\ \hline
Length & Number of words\\
Emoticon (Presence/Count)   & Boolean \& real-valued features indicating unigrams\\
Sentiment Ratio  & Positive and negative word ratios \\ \hline
\end{tabular}
\caption{Our Feature Set for Drunk-texting Prediction}
\label{tab:features}
\end{table*}
We use hashtag-based supervision to create our datasets, similar to tasks like emotion classification \cite{hashtags}. The tweets are downloaded using Twitter API (\url{https://dev.twitter.com/}). We remove non-Unicode characters, and eliminate tweets that contain hyperlinks\footnote{This is a rigid criterion, but we observe that tweets with hyperlinks are likely to be promotional in nature.} and also tweets that are shorter than 6 words in length. Finally, hashtags used to indicate drunk or sober tweets are removed so that they provide labels, but do not act as features. The dataset is available on request.  As a result, we create three datasets, each using a different strategy for sober tweets, as follows:
\begin{figure}[ht!]
\centering
\includegraphics[width=55mm, scale=0.7]{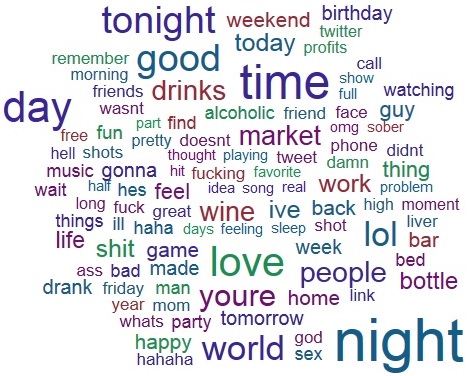}
\caption{Word cloud for drunk tweets}
\label{tab:topicsnondrunk}
\end{figure}
\begin{enumerate}
 \item \textbf{Dataset 1} (2435 drunk, 762 sober): We collect tweets that are marked as drunk and sober, using hashtags. Tweets containing hashtags \#drunk, \#drank and \#imdrunk are considered to be drunk tweets, while those with \#notdrunk, \#imnotdrunk and \#sober are considered to be sober tweets. 
 \item \textbf{Dataset 2} (2435 drunk, 5644 sober): The drunk tweets are downloaded using drunk hashtags, as above. The list of users who created these tweets is extracted. For the negative class, we download tweets by these users, which do not contain the hashtags that correspond to drunk tweets. 
\item \textbf{Dataset H} (193 drunk, 317 sober): A separate dataset is created where drunk tweets are downloaded using drunk hashtags, as above. The set of sober tweets is collected using both the approaches above. The resultant is the held-out test set \textit{Dataset-H that contains no tweets in common with Datasets 1 and 2}. 
\end{enumerate}

The drunk tweets for Datasets 1 and 2 are the same.  Figure~\ref{tab:topicsnondrunk} shows a word-cloud for these drunk tweets (with stop words and forms of the word `\textit{drunk}' removed), created using WordItOut\footnote{www.worditout.com}. The size of a word indicates its frequency. In addition to topical words such as `bar', `bottle' and `wine', the word-cloud shows sentiment words such as `love' or `damn', along with profane words.

Heuristics other than these hashtags could have been used for dataset creation. For example, timestamps were a good option to account for time at which a tweet was posted. However, this could not be used because user's local times was not available, since very few users had geolocation enabled.

\section{Feature Design}
\label{features}
The complete set of features is shown in Table~\ref{tab:features}. There are two sets of features: (a) N-gram features, and (b) Stylistic features. We use unigrams and bigrams as N-gram features- considering both presence and count.

Table~\ref{tab:features} shows the complete set of stylistic features of our prediction system. POS ratios are a set of features that record the proportion of each POS tag in the dataset (for example, the proportion of nouns/adjectives, etc.). The POS tags and named entity mentions are obtained from NLTK~\cite{nltk}. Discourse connectors are identified based on a manually created list. Spelling errors are identified using a spell checker by \newcite{enchant}. The repeated characters feature captures a situation in which a word contains a letter that is repeated three or more times, as in the case of \textit{happpy}. Since drunk-texting is often associated with emotional expression, we also incorporate a set of sentiment-based features. These features include: count/presence of emoticons and sentiment ratio. Sentiment ratio is the proportion of positive and negative words in the tweet. To determine positive and negative words, we use the sentiment lexicon in \newcite{mpqa}. To identify a more refined set of words that correspond to the two classes, we also estimated 20 topics for the dataset by estimating an LDA model~\cite{lda}. We then consider top 10 words per topic, for both classes. This results in 400 LDA-specific unigrams that are then used as features.

\begin{table}[h]
\begin{tabular}{|p{1.5cm}|p{0.6cm}|p{0.6cm}|p{0.6cm}|p{0.6cm}|p{0.6cm}|}
\hline
\multirow{2}{*}{} & \textbf{A (\%)} & \textbf{NP (\%)} & \textbf{PP (\%)} & \textbf{NR (\%)} & \textbf{PR (\%)} \\ \cline{1-6} 
\multicolumn{6}{|c|}{\textbf{Dataset 1}} \\ \hline 
%                 & \multicolumn{5}{c|}{\textbf{Na\"{i}ve Bayes}}                          \\ \hline
%N-gram           & 80.4        & 80.64       & 80.4        & 24.6        & 98.1        \\ \hline
%Stylistic          & 75.9       & 52.4        & 76.4        & 4.21        & 98.7        \\ \hline
%All  & 79.9       & 96.6        & 79.2        & 17.6        & 99.7        \\ \hline
 %                 & \multicolumn{5}{c|}{\textbf{SVM}}                                  \\ \hline
N-gram           & \textbf{85.5}       & 72.8        & 88.8        & 63.4        & 92.5        \\ \hline
Stylistic          & 75.6       & 32.5        & 76.2        & 3.2         & 98.6        \\ \hline
All   & 85.4       & 71.9        & 89.1        & 64.6        & 91.9        \\ \hline
\multicolumn{6}{|c|}{\textbf{Dataset 2}} \\ \hline 
 %                 & \multicolumn{5}{c|}{\textbf{Naive Bayes}}                          \\ \hline
%N-gram          & 77.4       & 79.5        & 68.6        & 91          & 45.6         \\ \hline
%Stylistic         & 70.4       & 71.1        & 55.1        & 97.2        & 7.9                 \\ \hline
%All  & 71.9       & 71.4        & 97.4        & 99.9        & 6.9             \\ \hline
%                  & \multicolumn{5}{c|}{\textbf{SVM}}                                  \\ \hline
N-gram          & 77.9       & 82.3        & 65.5        & 87.2        & 56.5     \\ \hline
Stylistic         & 70.3       & 70.8        & 56.7        & 97.9        & 6.01              \\ \hline
All & \textbf{78.1}       & 82.6        & 65.3        & 86.9        & 57.5                \\ \hline
\end{tabular}
\caption{Performance of our features on Datasets 1 and 2}
\label{tab:1}
\end{table}

\section{Evaluation}
Using the two sets of features, we train SVM classifiers~\cite{chang2011libsvm}\footnote{We also repeated all experiments for Na\"{i}ve Bayes. They do not perform as well as SVM, and have poor recall.}. We show the five-fold cross-validation performance of our features on Datasets 1 and 2, in Section~\ref{sec:cross}, and on Dataset H in Section~\ref{sec:heldout}. Section~\ref{sec:erroranal} presents an error analysis. \textit{Accuracy, positive/negative precision and positive/negative recall are shown as \textit{A}, \textit{PP/NP} and \textit{PR/NR} respectively. `Drunk' forms the positive class, while `Sober' forms the negative class.}
\begin{table}[h]
\begin{tabular}{|p{0.2cm}|p{1cm}|p{1cm}|}
\hline
\multicolumn{1}{|l|}{}              & \multicolumn{2}{c|}{\textbf{Top features}}                                          \\ \hline
\multicolumn{1}{|c|}{\textbf{\#}} & \multicolumn{1}{c|}{\textbf{Dataset 1}} & \multicolumn{1}{c|}{\textbf{Dataset 2}}   \\ \hline
\multicolumn{1}{|l|}{1}             & \multicolumn{1}{l|}{POS\_NOUN}          & \multicolumn{1}{l|}{Spelling\_error}   \\ \hline
\multicolumn{1}{|l|}{2}             & \multicolumn{1}{l|}{Capitalization}     & \multicolumn{1}{l|}{LDA\_drinking}        \\ \hline
\multicolumn{1}{|l|}{3}             & \multicolumn{1}{l|}{Spelling\_error} & \multicolumn{1}{l|}{POS\_NOUN}            \\ \hline
\multicolumn{1}{|l|}{4}             & \multicolumn{1}{l|}{POS\_PREPOSITION}   & \multicolumn{1}{l|}{Length}               \\ \hline
\multicolumn{1}{|l|}{5}             & \multicolumn{1}{l|}{Length}             & \multicolumn{1}{l|}{LDA\_tonight}         \\ \hline
\multicolumn{1}{|l|}{6}             & \multicolumn{1}{l|}{LDA\_Llife}         & \multicolumn{1}{l|}{Sentiment\_Ratio}      \\ \hline
\multicolumn{1}{|l|}{7}             & \multicolumn{1}{l|}{POS\_VERB}          & \multicolumn{1}{l|}{Char\_repeat} \\ \hline
\multicolumn{1}{|l|}{8}             & \multicolumn{1}{l|}{LDA\_today}         & \multicolumn{1}{l|}{LDA\_today}           \\ \hline
\multicolumn{1}{|l|}{9}             & \multicolumn{1}{l|}{POS\_ADV}           & \multicolumn{1}{l|}{LDA\_drunken}         \\ \hline
\multicolumn{1}{|l|}{10}                                   & Sentiment\_Ratio                         & LDA\_lmao      \\ \hline                          
\end{tabular}
\caption{Top stylistic features for Datasets 1 and 2 obtained using Chi-squared test-based ranking}
\label{tab:topfeatures}
\end{table}

\subsection{Performance for Datasets 1 and 2}
\label{sec:cross}
Table ~\ref{tab:1} shows the performance for five-fold cross-validation for Datasets 1 and 2. In case of Dataset 1, we observe that N-gram features achieve an accuracy of 85.5\%. We see that our stylistic features alone exhibit degraded performance, with an accuracy of 75.6\%, in the case of Dataset 1. Table~\ref{tab:topfeatures} shows top stylistic features, when trained on the two datasets. Spelling errors, POS ratios for nouns (POS\_NOUN)\footnote{POS ratios for nouns, adjectives and adverbs were nearly similar in drunk and sober tweets - with the maximum difference being 0.03\%}, length and sentiment ratios appear in both lists, in addition to LDA-based unigrams. However, negative recall reduces to a mere 3.2\%. This degradation implies that our features capture a subset of drunk tweets and that there are properties of drunk tweets that may be more subtle. 
When both N-gram and stylistic features are used, there is negligible improvement. The accuracy for Dataset 2 increases from 77.9\% to 78.1\%. Precision/Recall metrics do not change significantly either. The best accuracy of our classifier is 78.1\% for all features, and 75.6\% for stylistic features. This shows that text-based clues can indeed be used for drunk-texting prediction.

\begin{table}[h]
\centering
\begin{tabular}{|c|c|c|c|}
\hline
   & A1            & A2            & A3      \\ \hline   %& H             \\ \hline
A1 & \textbf{-}    & 0.42 & 0.36 \\ \hline %& \textit{0.35} \\ \hline
A2 & 0.42 & \textbf{-}    & 0.30 \\ \hline %& \textit{0.32} \\ \hline
A3 & 0.36    & 0.30    & \textbf{-}  \\ \hline %& \textit{0.187}    \\ \hline
%H  & \textit{0.35} & \textit{0.32} & \textit{0.187} \\ \hline %& \textbf{-}    \\ \hline
\end{tabular}
\caption{Cohen's Kappa for three annotators (A1-A3)}
\label{tab:iaa}
\end{table}
\begin{table}[ht!]
\centering
\begin{tabular}{|p{2.2cm}|p{0.6cm}|p{0.6cm}|p{0.6cm}|p{0.6cm}|p{0.6cm}|}
\hline
\multirow{2}{*}{} & \textbf{A (\%)} & \textbf{NP (\%)} & \textbf{PP (\%)} & \textbf{NR (\%)} & \textbf{PR (\%)} \\ \cline{1-6} 
%\multicolumn{6}{|c|}{\textbf{Dataset H}} \\ \hline 
Annotators & 68.8 & 71.7 & 61.7 & 83.9 & 43.5 \\ \hline
   \textbf{Training Dataset}         & \multicolumn{5}{c|}{\textbf{Our classifiers}}                          \\ \hline
%Dataset 1 & 39 &	100	& 38 &	1 &	100 \\ \hline
% Dataset 2 & 62.7	& 62.5 &	80	& 99.6 & 2.1\\ \hline
 %                 & \multicolumn{5}{c|}{\textbf{SVM}}                                  \\ \hline
Dataset 1 & 47.3 &	70	& 40 &	26 &	81 \\ \hline
Dataset 2 & 	64 &	70 &	53 &	72 &	50 \\ \hline
 \end{tabular}
\caption{Performance of human evaluators and our classifiers (trained on all features), for Dataset-H as the test set}
\label{tab:2}
\end{table}
\subsection{Performance for Held-out Dataset H}
\label{sec:heldout} 
Using held-out dataset H, we evaluate how our system performs in comparison to humans. Three annotators, A1-A3, mark each tweet in the Dataset H as drunk or sober. Table ~\ref{tab:iaa} shows a moderate agreement between our annotators (for example, it is 0.42 for A1 and A2). Table ~\ref{tab:2} compares our classifier with humans. Our human annotators perform the task with an average accuracy of 68.8\%, while our classifier (with all features) trained on Dataset 2 reaches 64\%. The classifier trained on Dataset 2 is better than which is trained on Dataset 1.

\subsection{Error Analysis}
\label{sec:erroranal}
Some categories of errors that occur are:
\begin{enumerate}
\item \textbf{Incorrect hashtag supervision}: The tweet `\textit{Can't believe I lost my bag last night, literally had everything in! Thanks god the bar man found it}' was marked with`\#Drunk'. However, this tweet is not likely to be a drunk tweet, but describes a drunk episode in retrospective. Our classifier predicts it as sober.
\item \textbf{Seemingly sober tweets}: Human annotators as well as our classifier could not identify whether `\textit{Will you take her on a date? But really she does like you}' was drunk, although the author of the tweet had marked it so. This example also highlights the difficulty of drunk-texting prediction.
\item \textbf{Pragmatic difficulty}: The tweet `\textit{National dress of Ireland is one's one vomit.. my family is lovely}' was correctly identified by our human annotators as a drunk tweet. This tweet contains an element of humour and topic change, but our classifier could not capture it.
\end{enumerate}

\section{Conclusion \& Future Work}
In this paper, we introduce automatic drunk-texting prediction as the task of predicting a tweet as drunk or sober. First, we justify the need for drunk-texting prediction as means of identifying risky social behavior arising out of alcohol abuse, and the need to build tools that avoid privacy leaks due to drunk-texting. We then highlight the challenges of drunk-texting prediction: one of the challenges is selection of negative examples (sober tweets). Using hashtag-based supervision, we create three datasets annotated with drunk or sober labels. We then present SVM-based classifiers which use two sets of features: N-gram and stylistic features. Our drunk prediction system obtains a best accuracy of 78.1\%. We observe that our stylistic features add negligible value to N-gram features.  We use our heldout dataset to compare how our system performs against human annotators. While human annotators achieve an accuracy of 68.8\%, our system reaches reasonably close and performs with a best accuracy of 64\%. 

Our analysis of the task and experimental findings make a case for drunk-texting prediction as a useful and feasible NLP application.

\bibliographystyle{acl}
\bibliography{acl2015}
\end{document}